\documentclass{article}
\pdfoutput=1
\usepackage[preprint, nonatbib]{nips_2018}

\usepackage{float}

\usepackage[square,numbers]{natbib}
\usepackage[utf8]{inputenc} % allow utf-8 input
\usepackage[T1]{fontenc}    % use 8-bit T1 fonts
\usepackage{hyperref}       % hyperlinks
\usepackage{url}            % simple URL typesetting
\usepackage{booktabs}       % professional-quality tables
\usepackage{amsfonts}       % blackboard math symbols
\usepackage{nicefrac}       % compact symbols for 1/2, etc.
\usepackage{microtype}      % microtypography
\usepackage{graphicx}
\usepackage{tabularx}
\usepackage{hyperref}

\usepackage{algorithm}
\usepackage{algorithmic}
\usepackage{amsmath,amsfonts}
\usepackage{amssymb}
\usepackage{bbding}
\usepackage{booktabs}
\usepackage{multirow}
\usepackage{subfig}
\usepackage{lineno}
\usepackage{color}
\usepackage{graphicx}
\usepackage{soul} % 导入 soul 包
\usepackage{color, xcolor}

\title{Bootstrapping Audio-Visual Segmentation by Strengthening Audio Cues}

\author{
  Tianxiang Chen \\
  USTC, China \\
  Alibaba Group, China \\
  \texttt{txchen@mail.ustc.edu.cn} \\
\And
 Zhentao Tan \\
  USTC, China \\
  Alibaba Group, China \\
  \texttt{tzt@mail.ustc.edu.cn} \\
\And
Tao Gong \\
  USTC, China \\
  \texttt{gt950513@mail.ustc.edu.cn} \\
\And
Qi Chu \\
  USTC, China \\
  \texttt{qchu@ustc.edu.cn} \\
\And
Yue Wu \\
  Alibaba Group, China \\
  \texttt{matthew.wy@alibaba-inc.com} \\
\And
Bin Liu \\
USTC, China \\
  \texttt{flowice@ustc.edu.cn} \\
\And
Le Lu \\
  Alibaba Group, China \\
  \texttt{tiger.lelu@gmail.com} \\
\And
Jieping Ye \\
  Alibaba Group, China \\
  \texttt{yejieping.ye@alibaba-inc.com} \\
\And
Nenghai Yu \\
  USTC, China \\
  \texttt{ynh@ustc.edu.cn} \\
}

\begin{document}
\maketitle

\begin{abstract}
How to effectively interact audio with vision has garnered considerable interest within the multi-modality research field. Recently, a novel audio-visual 
segmentation (AVS) task has been proposed, aiming to segment the sounding objects in video frames under the guidance of audio cues. However, most existing AVS methods are hindered by a modality imbalance where the visual features tend to dominate those of the audio modality, due to a unidirectional and insufficient integration of audio cues. This imbalance skews the feature representation towards the visual aspect, impeding the learning of joint audio-visual representations and potentially causing segmentation inaccuracies. To address this issue, we propose AVSAC. Our approach features a Bidirectional Audio-Visual Decoder (BAVD) with integrated bidirectional bridges, enhancing audio cues and fostering continuous interplay between audio and visual modalities. This bidirectional interaction narrows the modality imbalance, facilitating more effective learning of integrated audio-visual representations. Additionally, we present a strategy for audio-visual frame-wise synchrony as fine-grained guidance of BAVD. This strategy enhances the share of auditory components in visual features, contributing to a more balanced audio-visual representation learning. Extensive experiments show that our method attains new benchmarks in AVS performance.
\end{abstract}

\section{Introduction}
\label{submission}

Audio and vision are two closely intertwined modalities that play an indispensable role in our perception of the real world. Recently, a novel audio-visual segmentation (AVS) \cite{zhou2022audio} task has been proposed, which aims at segmenting the sounding objects from video frames corresponding to a given audio. The task has rapidly raised wide interest from many researchers since being proposed, while there are still some issues remaining to be solved. One of the biggest challenges is that this task requires both accurate location of the audible frames and precise delineation of the shapes of the sounding objects \cite{zhou2022audio, zhou2023audio}. This necessitates reasoning and adequate interaction between audio and vision modalities for a deeper understanding of the scenarios, which renders many present audio-visual task-related methods \cite{qian2020multiple,chen2021localizing} unsuitable for AVS. Therefore, it is necessary to tailor a new method for AVS.

AVSBench \cite{zhou2022audio} is the first baseline method for AVS, but its convolutional structure limits its performance due to limited receptive fields \cite{gao2023avsegformer}. As a remedy, transformers have recently been introduced to AVS network designs \cite{gao2023avsegformer, liu2023audio, liu2023bavs} by using the attention mechanism to model the audio-visual relationship. The attention mechanism can highlight the most relevant video frame regions for each audio input by aggregating the visual features according to the audio-guided attention weights. 

\begin{figure*}
    \centering
    \includegraphics[width=0.97\textwidth,height=0.35\textwidth]{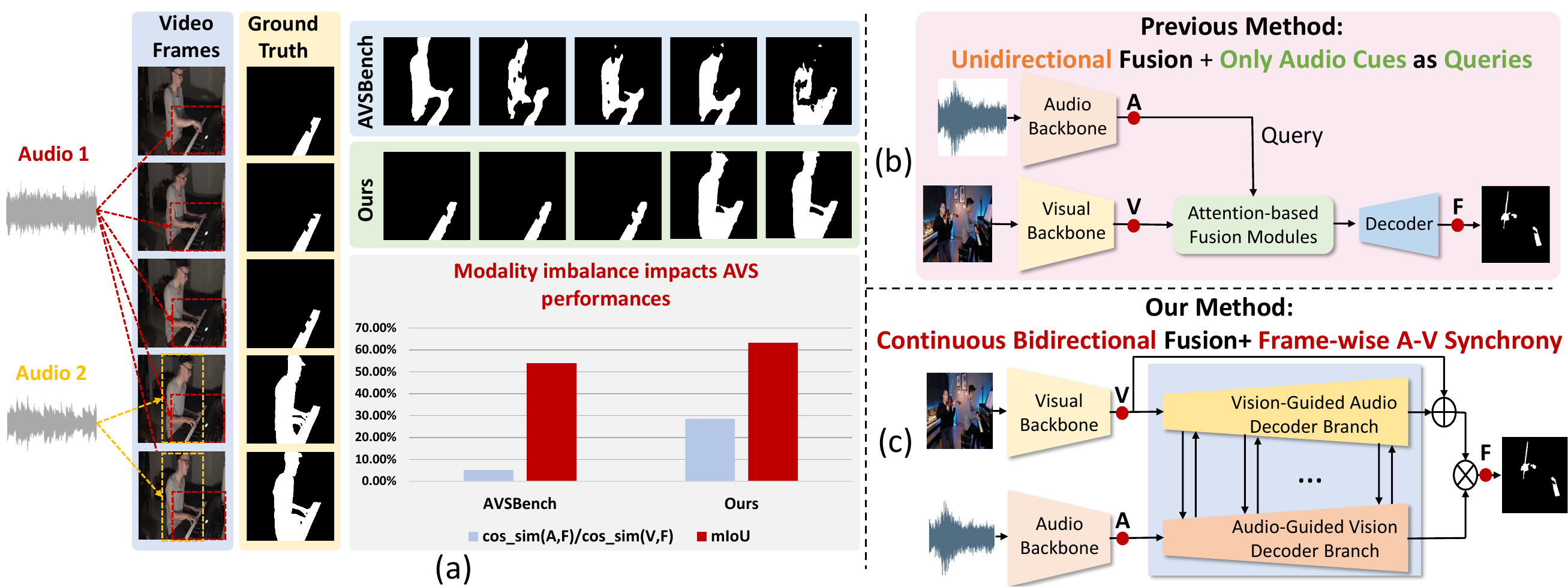}
    \caption{(a) Modality imbalance in the AVS task, where visual features tend to overshadow audio cues and impact the AVS result. %We mark in red dots in (b) and (c) the final output feature before the segmentation head as F, the audio backbone output as A, and the visual backbone output as V. 
    We adopt the proportion of audio and visual feature components included in the final feature to measure the audio-visual imbalance degree, combining with visual results for analysis. %We find that modality imbalance actually harms the AVS fineness as some audio cues are overwhelmed so much that cannot provide effective guidance. 
    The reasons for such imbalance are that (b) the audio-visual fusion mode of present methods \cite{zhou2022audio,gao2023avsegformer} is unidirectional and insufficient, with only audio cues as queries. 
    (c) Our AVSAC can relieve modality imbalance and features a paralleled decoder structure with multiple bidirectional bridges linked within to strengthen audio cues through bidirectional and continuous audio-visual interaction. Also, we introduce audio-visual frame-wise synchrony to foster the integration of audio components into visual features for further modality imbalance alleviation.}\label{compare}
\end{figure*}

Despite some improvements, present AVS methods still suffer a modality imbalance issue: the network tends to focus more on the visual information while neglecting some effective audio guidance. As illustrated in Figure ~\ref{compare} (a), We assess the degree of audio-visual imbalance by gauging the ratio of auditory to visual components in the final feature and find that the auditory component only constitutes a minor fraction in a present method \cite{zhou2022audio}. Notably, modality imbalance generally exists in multi-modality fields \cite{han2022trusted, peng2022balanced, zhou2020improving,liu2023multi}. But in the AVS field, such imbalance is deteriorated due to the unidirectional and deficient fusion (only audio cues serve as queries) of present AVS methods, in contrast to the far more exploited vision information \cite{zhou2022audio,gao2023avsegformer,liu2023audio, liu2023bavs}. This is concluded in Figure ~\ref{compare} (b), where the inadequate audio cue either repeats itself to the same size as the visual feature for cross attention \cite{zhou2022audio} or serves as a query for channel attention \cite{gao2023avsegformer} in the fusion modules to produce what we define the audio-guided vision features (AGV). %This implies a modality imbalance issue: 
%the network tend to focus more on the visual information while neglecting some effective audio guidance. %because the audio cue is not as information-rich as the visual feature. Present methods may easily cause some audio cues to fade away. 
The worsened imbalance harms audio-visual representation learning, as some audio cues are overwhelmed so much that cannot provide effective guidance to visual segmentation. Thus, it's intuitive to derive vision-guided audio features (VGA) to balance with AGV by aggregating the audio features together according to the vision-guided attention weights. However, both AGV and VGA are single-modal features that only represent part of the multi-modal information. For instance, VGA is a set of audio features that describes pixels but does not preserve the inherent visual feature of each pixel itself. We argue that a holistic understanding of audio-visual representation can be obtained through bidirectional modality interaction.

%AVSBench \citep{zhou2022audio} fuses audio and visual modalities in an unidirectional way instead of bilateral interaction. AVSegFormer \citep{gao2023avsegformer} (also other transformer-based AVS methods \citep{liu2023audio, liu2023bavs}) only feed the last-layer visual representations of the encoder to the audio decoder, causing heavy visual information loss.  Present AVS methods only use the audio feature as query to generate the attention weights in fusion modules and get what we define audio-guided vision features (AGV). Thus, the output features tend to be dominated by visual representations while the audio information may fade away during the information propagation along the network, which restricts audio-vision representation learning and brings uncertainty to the mask prediction because the audio representation is also essential to AVS. 

Motivated by this, we devise AVSAC, featuring a dual-tower decoder architecture that employs bidirectional bridges to connect each layer. It integrates AGV and VGA into one structure, as shown in Figure ~\ref{compare} (c). The vision-guided decoder branch processes and outputs the VGA, while the audio-guided decoder branch processes and outputs AGV. The bidirectional bridges enable continuous and in-depth interaction between the two modalities to boost the significance of audio cues. In addition, we incorporate audio-visual frame-wise synchrony (AVFS) into our framework by introducing an AVFS loss to further relieve modality imbalances, fostering a more balanced and integrated audio-visual learning process. The AVFS loss ensures that visual features can learn from audio cues, thereby lifting the ratio of audio components within the visual features. Our method significantly lifts the audio component ratio compared with other methods according to Figure ~\ref{compare} (a). %AVTS maintains the consistency of audio-visual signals over sequential frames, enhancing the overall accuracy of the segmentation task.

%In addition, since most AVS methods use the GT mask as the only supervision, no effective feedback can be given to prevent the network from learning a certain data bias \cite{chen2023closer} that enables the network to “guess” the correct target even if the audio cue has been lost. We hence propose the Audio Feature Reconstruction (AFR) to protect network from harmful data bias and guarantee the existence of audio cues. AFR loss is then introduced to supervise network training.

Our main \textbf{contributions} can be summarized as three-fold: 

\begin{itemize} 
\item We propose a Bidirectional Audio-Visual Decoder (BAVD) to strengthen audio cues through continuous and in-depth audio-visual interaction. %Page limit: The main body of the paper has to be fitted to 8 pages, excluding references and appendices; the space for the latter two is not limited. For the final version of the paper, authors can add one extra page to the main body.
\item We propose a new audio-visual frame-wise synchrony loss function to enable visual signals to learn from audio cues on a per-frame basis, enriching the auditory component within visual features.
\item Our method achieves new state-of-the-art performances on three sub-tasks of the AVS benchmark.
\end{itemize}

\begin{figure*}
    \centering
    \includegraphics[width=0.99\textwidth,height=0.45\textwidth]{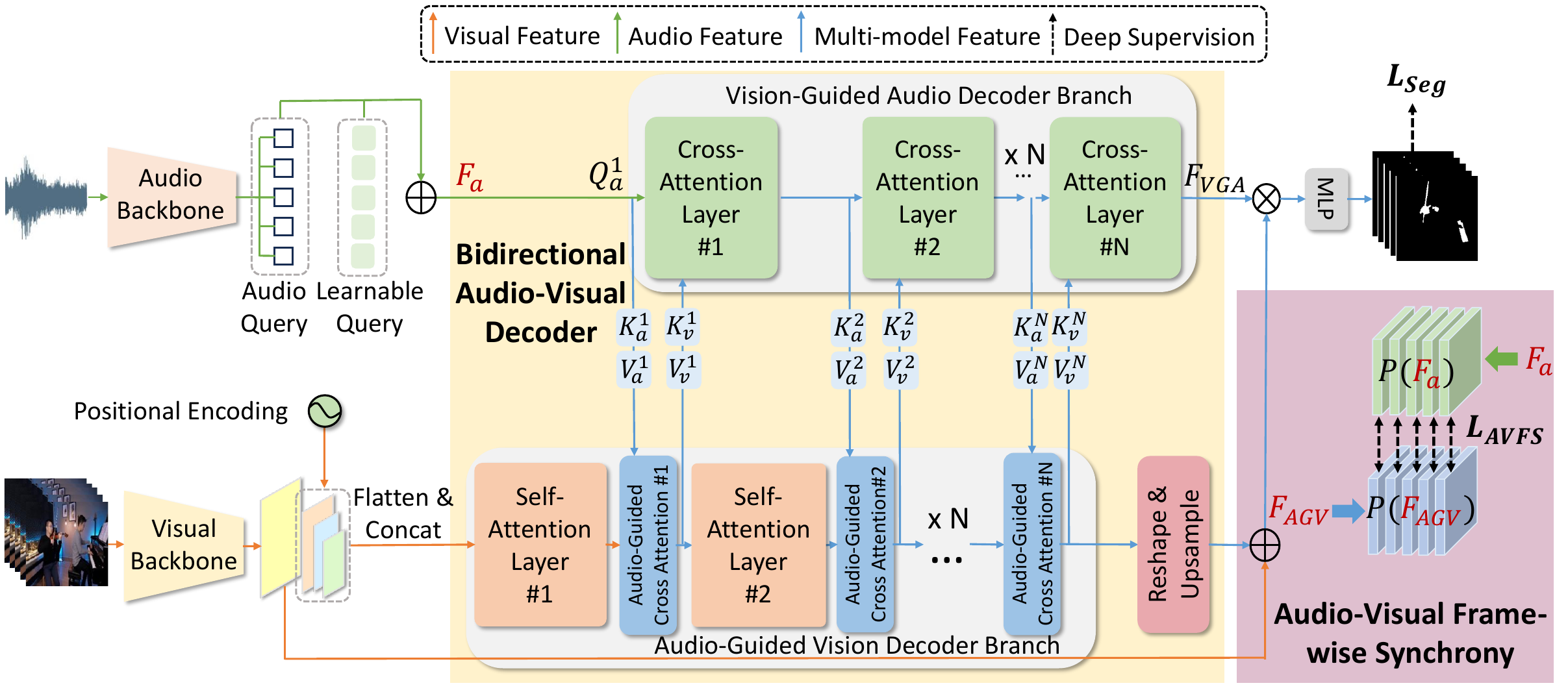}
    \caption{Overall architecture of AVSAC. We propose two key components in this framework: (1) Bidirectional Audio-Visual Decoder (BAVD), enabling the model to strengthen audio cues by consistently balancing AGV with VGA through continuous and in-depth bilateral modality interaction. (2) Audio-Visual Frame-wise Synchrony (AVFS) module is proposed as a more fine-grained (frame-wise) guidance to our BAVD. It can help exploit auditory components to visual features to increase the importance of audio cues.}\label{overview}
\end{figure*}

\section{Related Work}

\subsection{Audio-Visual Segmentation}
Audio-visual segmentation (AVS) aims to predict pixel-wise masks of the sounding objects in a video sequence given audio cues. To tackle this issue, \citep{zhou2022audio, zhou2023audio} proposed the first audio-visual segmentation benchmark with pixel-level annotations and devised AVSBench based on the temporal audio-visual fusion with audio features as query and visual features as keys and values. However, the relatively limited receptive field of convolutions restricts the AVS performance. As a remedy, AVSegFormer \citep{gao2023avsegformer} was later proposed based on the vision transformer and achieved good AVS performances. Following \citep{gao2023avsegformer}, a series of ViT-based AVS networks has also been proposed\citep{liu2023audio, liu2023bavs,li2023catr}. Recently \citep{mao2023contrastive} has introduced a diffusion model to AVS and attained good results. However, these methods do not pay enough attention to audio cues as to visual features, in that the fusion of audio features is relatively insufficient and unidirectional, while the vision information is more sufficiently exploited by sending it to every decoder layer. Therefore, the audio-visual imbalance problem worsened, making some effective audio guidance easily fade away. To solve this, we propose to continuously strengthen audio cues to reach modality balance.

%Then these methods use the scanty audio feature as a query to generate the attention weights in fusion modules to guide the vision features. This inevitably triggers modality imbalance--the network will tend to focus more on the vision information instead of the little bits of audio cues, making the audio information easily fade away as the information propagates along the network. Therefore, the audio response will not be enough to help the network accurately distinguish what is the sounding object from the complex background. 

%To solve the modality imbalance challenge, the intuitive is to balance the abundant visual features with audio information. To achieve this we need to guarantee that the audio cues cannot fade away during information propagation, so we choose a two-tower decoder AVS network structure with multiple bidirectional bridges linked between each layer for continuous and in-depth audio-visual information interaction. During this process, the audio cues can be continuously boosted for modality balance.  

\subsection{Vision Transformer}
Drawing inspirations from Deformable DETR \cite{zhu2020deformable} and DINO \cite{zhang2022dino}, AVSegFormer \cite{gao2023avsegformer} leveraged ViT to extract global dependencies. The remarkable performance of AVSegFormer \cite{gao2023avsegformer} has inspired more researchers to apply ViT to AVS task \cite{liu2023audio, liu2023bavs,li2023catr}. Nevertheless, the integration of audio features in these approaches is flawed because it is unidirectional and inadequate, while the vision information is more sufficiently exploited by sending it to every decoder layer for attention operations. The discrepancy inevitably triggers a modality imbalance, with the network disproportionately favoring visual features over the scant audio cues, resulting in the audio information gradually being overshadowed. To solve this, we devise AVSAC, grounded in the ViT framework, to enhance the disadvantaged audio information so as to balance with visual features.

The most related work of our paper is \cite{liu2023multi}, which uses ViT and also notices the modality imbalance problem but in a language-vision task. However, its feature extraction process only considers vision modality while we involve both audio and vision modalities. Also, the loss functions for network guidance are completely different.

\section{Method}
\subsection{Overview}

Figure~\ref{overview} presents a schematic overview of our AVSAC, showcasing a dual-tower decoder architecture that is interconnected through a series of bidirectional bridges. The network accepts the audio and video frames as inputs. Following previous works \citep{zhou2022audio,gao2023avsegformer} in feature extraction, we first extract two sets of features: the visual feature $F=\{F_{1},F_{2},F_{3},F_{4}\}$ from a CNN backbone (\citep{he2016deep,wang2022pvt}), where $F_{i}\in \mathbb{R}^{T \times 256 \times \frac{H}{2^{i+1}} \times \frac{W}{2^{i+1}}}, i\in \{1,2,3,4\}$; and the audio feature $F_{a}\in \mathbb{R}^{T \times 1 \times 256}$ from an audio backbone \citep{hershey2017cnn}, where $T$ is the number of frames. Specifically, $F_{a}$ is the integration of the learnable query and the audio query. %The query integration can boost our model’s adaptability for multiple AVS tasks and datasets. 
We collect $F_{2},F_{3},F_{4}$ and then flatten and concatenate them as $F_{vis}\in \mathbb{R}^{T \times L \times 256}$. Then $F_{a}$ and $F_{vis}$ are fed into corresponding decoders for multi-model interaction and output binary segmentation masks. Our AVSAC has two key components: (1) a Bidirectional Audio-Visual Decoder (BAVD) and (2) an Audio-Visual Frame-wise Synchrony (AVFS) loss. We detail each component of our framework in the following.

\subsection{Bidirectional Audio-Visual Decoder (BAVD)}
As mentioned above, most present AVS methods use the naive attention mechanism to process multi-modal information but always use audio features as query and visual features as key and value through channel attention or repeating audio features to the same size as visual features. In this way, the audio features only attend the generation of attention weights that indicate the significance of each area in the visual feature but do not directly involve in the output, so that the auditory component within the output is relatively low, so we define it as the audio-guided vision feature (AGV). Even worse is that the AGV is then sent to the successive transformer decoder. As a result, visual features take the dominance while audio features dramatically get lost in the decoder. This type of modality fusion is inefficient, the visual features prone to overshadow the auditory ones and worsen the modality imbalance.

To relieve modality imbalance, we hope to equally treat features from both modalities and interact with each other. Based on this idea, we propose a Bidirectional Audio-Visual Decoder (BAVD) that consists of two paralleled interactive decoders with bidirectional bridges, as shown in Figure ~\ref{overview}. The two-tower decoder structure includes an audio-guided decoder branch and a vision-guided decoder branch, responsible for extracting AGV and VGA respectively. In this way, both AGV and VGA can be consistently extracted and mutually balance each other as the network goes deep, which significantly relieves modality imbalance. To obtain the final segmentation mask prediction, we multiply the AGV feature output $F_{AGV}$ obtained from the audio-guided vision decoder branch with the VGA feature output $F_{VGA}$ from the vision-guided audio decoder branch. Then we utilize an MLP to integrate different channels, followed by a residual connection to ensure that the fusion of audio representation does not lead to aggressive loss of visual representation. Finally, the segmentation mask $M$ can be obtained through a linear layer. The process be expressed as Equation ~\ref{final}.
\begin{equation}\label{final}
M=Linear(F_{AGV}+MLP(F_{AGV} \cdot F_{VGA} ))
\end{equation}
where $MLP(.)$ and $Linear(.)$ denote the MLP process and the linear layer, respectively. In the following, we will introduce the two decoder branches in our BAVD in detail on how to obtain $F_{VGA}$ and $F_{AGV}$.

\begin{figure}
    \centering
    \includegraphics[width=0.98\textwidth,height=0.30\textwidth]{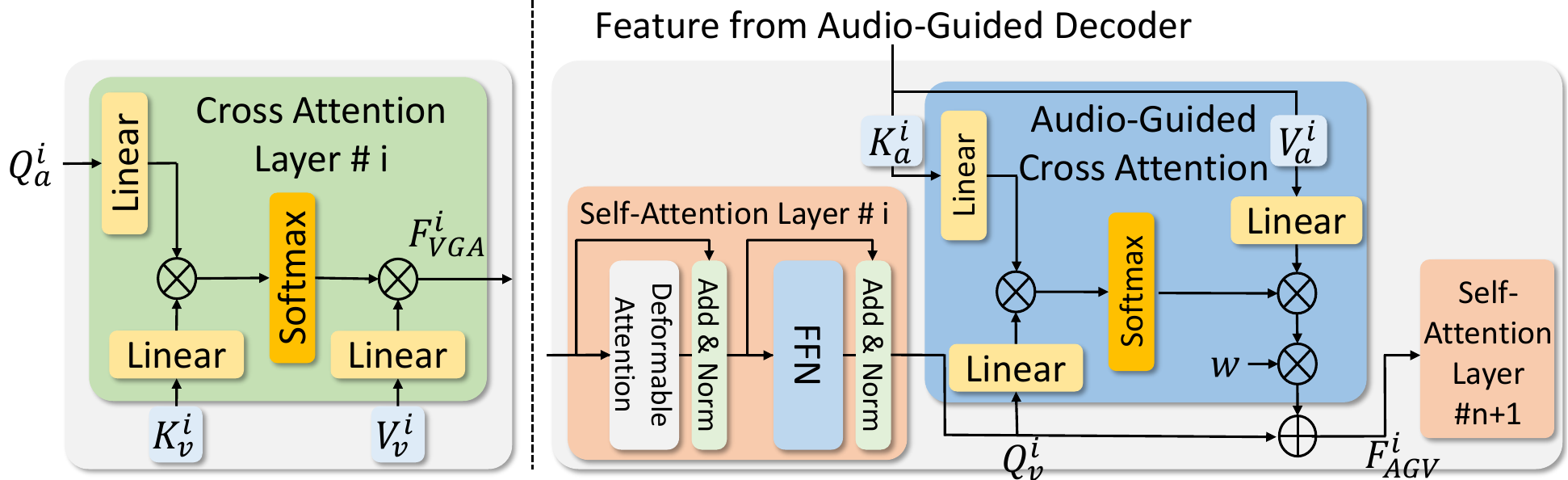}
    \caption{Illustration of the Internal module structure of Bidirectional Audio-Visual Vision Decoder (BAVD).}\label{bavd}
\end{figure}

\subsubsection{audio-guided vision (AGV) decoder branch}
We adopt the deformable attention block following \citep{zhu2020deformable} as the basic structure of the self-attention layer in the AGV decoder branch to generate visual queries for audio-guided cross attention in the next step, as shown in the right part of Figure ~\ref{bavd}. We build multiple bidirectional bridges between each layer of the two decoders for bidirectional modality interaction. Specifically, the inputs of i-th layer of the VGA decoder branch are injected into the i-th layer of the AGV decoder branch as key ($K_{a}^{i}$) and value ($V_{a}^{i}$) for multi-modality interaction based on the cross-attention operation. The expression of audio-guided cross attention on the i-th layer is written as Equation~\ref{cross1}.
\begin{equation}\label{cross1}
F_{AGV}^{i}=F_{v}^{i}+Softmax(\frac{Q_{v}^{i}K_{a}^{i^T}}{\sqrt{d_{K}}})V_{a}^{i}w
\end{equation}
where $F_{AGV}^{i} \in \mathbb{R}^{T \times L \times 256}$ is the i-th audio-guided cross attention output, $K_{a}^{i},V_{a}^{i}$ are the key and value sent through the bridge from the i-th layer of the VGA decoder branch and $Q_{v}^{i}$ is the query of the i-th layer of the AGV decoder branch. $F_{v}^{i}$ is the i-th layer self attention output. $w$ is a learnable weight that enables the AGV decoder branch to determine how much information it needs and enhance the adaptability of audio cues to the joint representation. 

\subsubsection{vision-guided audio (VGA) decoder branch}
The basic operation in our VGA decoder branch is the cross-attention operation, as shown in the left part of Figure~\ref{bavd}. We do not follow the same structure as the AGV decoder branch as we delete the self-attention layer because we find that the audio representation is not as information-rich as the visual representation, and additional self-attention layers only bring increased parameters but dropped performance. The experimental result is recorded in our ablation study in Table~\ref{ab2}. The reason is that excessive parameters may cause ambiguity in audio representations to harm the performance. Therefore we simplify our VGA decoder branch to only cross attention layers. The expression of the cross attention on the i-th layer is written as Equation~\ref{cross2}.

\begin{equation}\label{cross2}
F_{VGA}^{i}=Softmax(\frac{Q_{a}^{i}K_{v}^{i^T}}{\sqrt{d_{k}}})V_{v}^{i}, \quad Q_{a}^{i}=F_{VGA}^{i-1}
\end{equation}

where $F_{VGA}^{i} \in \mathbb{R}^{T \times 300 \times 256}$ is the i-th layer cross attention output. $Q_{a}^{i}$ is the query of the i-th layer of the VGA decoder branch, which is also the former layer cross-attention output. and $K_{v}^{i},V_{v}^{i}$ are the key and value come from the i-th layer outputs of the AGV decoder branch.

\subsection{Audio-Visual Frame-wise Synchrony (AVFS)} 
% Audio features are usually more sparse and abstract than visual features and easily get dominated by visual features. By emphasizing the synchronization between audio and visual features, the model can better capture the correlation between both two modalities, thereby enhancing the importance of audio features in AVS tasks and reducing the dominant effect of visual information.

% Given the audio feature $F_{a}$ extracted from the audio backbone and the audio-guided vision feature $F_{AGV}$, we extend and reshape $F_{a}$ to match the shape of $F_{AGV}$, flatten the two features and then apply the softmax function to convert the two features into probability distributions. Next, we adopt the Jensen-Shannon (JS) divergence as a constraint to measure the similarity between the audio and visual features. By minimizing the JS divergence, we hope to encourage the  distributions of the two modalities to be as close as
% possible in a self-supervised manner, thus promoting their synchrony during the AVS process. The S3 loss can be
% formulated as:
% \begin{equation}
% \begin{aligned}
% &\mathcal{L}_{S3}=\frac{1}{2}KL(F_{a}||M)+\frac{1}{2}KL(F_{VGA}||M),\\
% &M=\frac{1}{2}(F_{a}+F_{VGA}), KL(\mathcal{P}||\mathcal{Q})=\sum \mathcal{P}log\frac{\mathcal{P}(t)}{\mathcal{Q}(t)}\\
% \end{aligned}
% \end{equation} 

Audio cues are usually more sparse and abstract than visual features and easily get dominated by visual features. The above proposed BAVD tackles this imbalance through network structural modifications, but lacks a certain constraint to guide the learning process towards achieving modality balance. Thereby, we introduce the Audio-Visual Frame-wise Synchrony (AVFS) loss. The loss can emphasize the frame-wise synchronization between audio and visual feature pairs, thereby enabling the visual feature to learn from audio cues to enhance its auditory component. AVFS can be regarded as a fine-grained guidance to enhance the importance of audio features. 

Given the audio feature $F_{a}$, which is the audio input of our VGA decoder branch, and the audio-guided vision feature $F_{AGV}$, we extend and reshape $F_{a}$ to match the shape of $F_{AGV}$, then apply the softmax function to convert the two features into probability distributions. To enable $F_{AGV}$ to exploit auditory parts from $F_{a}$, we flatten the two distribution features to the shape of $\mathbb{R}^{T \times 256 \times (128\times128)}$ and adopt the Kullback-Leibler (KL) divergence as a constraint to measure the similarity between the audio and visual features. Notably, the KL divergence is constrained on each corresponding frame pair of the audio and visual distribution features. The AVFS loss can be formulated as:

\begin{equation}
\mathcal{L}_{AVFS} = \frac{1}{T} \sum_{t=1}^{T} P(F_{a})_{t,:,:} \cdot \log\left(\frac{P(F_{a})_{t,:,:}}{P(F_{AGV})_{t,:,:}}\right)\\
\end{equation} 

where $P(.)$ denotes the above-mentioned operations to transform modality features into probability distribution features, and $T$ stands for the number of frames. Our proposed AVFS strategy does not participate in mask prediction and is computationally free during inference. It can serve as a plug-in module to any existing AVS methods.

\subsection{Loss Function}
The total loss function $\mathcal{L}$ consists of two parts: one is the loss for shape-aware segmentation ($\mathcal{L}_{Seg}$) that includes binary Focal loss $\mathcal{L}_{focal}$ \cite{lin2017focal} and Dice loss $\mathcal{L}_{dice}$ \cite{milletari2016v}, and the other is for audio-visual temporal synchrony ($\mathcal{L}_{AVFS}$). We define $\mathcal{L}_{Seg}$ =$  \mathcal{L}_{focal}$ + $\mathcal{L}_{dice}$. We use $\mathcal{L}_{Seg}$ to supervise the final MLP output and $\mathcal{L}_{AVFS}$ to supervise the original audio input of BAVD ($F_{a}$) and the output audio-guided visual feature $F_{AGV}$. The total loss function can be written as

\begin{equation}
\mathcal{L}(\hat{y},y,F_{a},F_{AGV} )= \mathcal{L}_{Seg}(\hat{y},y) + \mathcal{L}_{AVFS} = \mathcal{L}_{dice}(\hat{y},y)+ \mathcal{L}_{focal}(\hat{y},y)+ \mathcal{L}_{AVFS}
\end{equation} 

The whole framework is trained by minimizing the segmentation loss function between $\hat{y}$ and ground-truth $y$ and the AVFS loss between $F_{a}$ and $F_{AGV}$. %Our proposed audio-visual temporal synchrony strategy does not participate in mask prediction and is computationally free during inference. It can serve as a plug-in module to any existing AVS methods.

\begin{figure*}
    \centering
    \includegraphics[width=0.99\textwidth,height=0.38\textwidth]{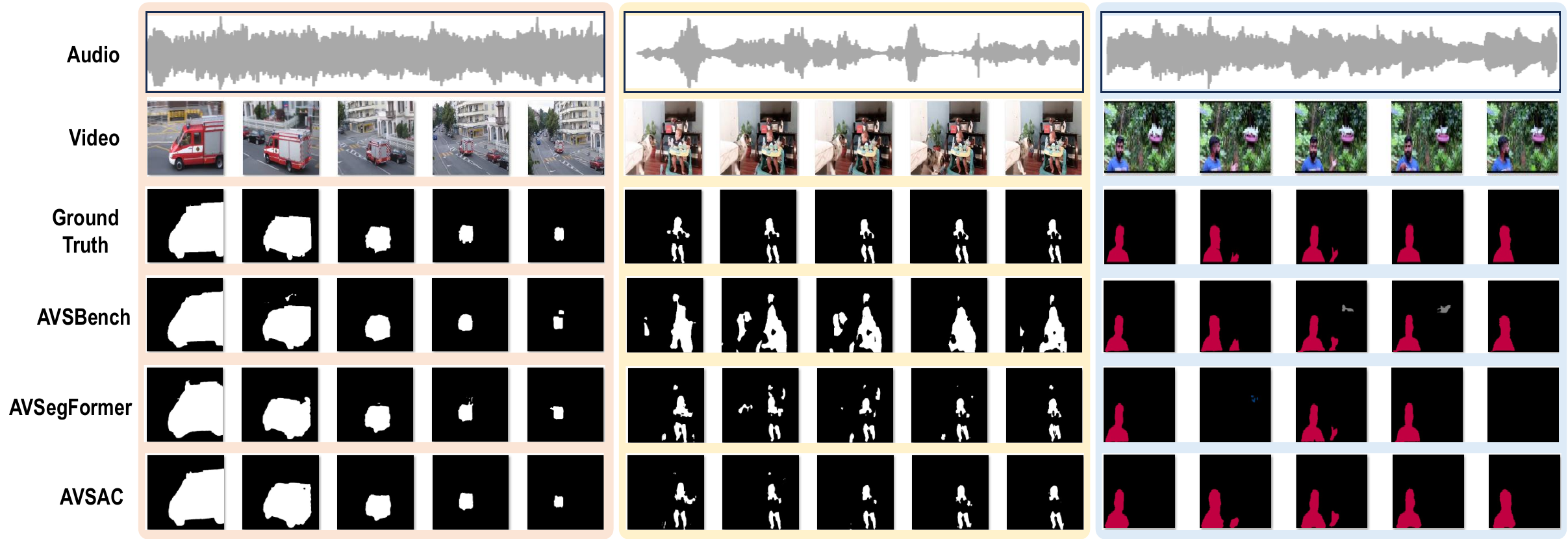}
    \caption{Qualitative examples of the AVSBench, AVSegFormer, and our AVSAC framework. The left part shows the video frame outputs of the S4 setting, the middle part offers the video frame outputs of the MS3 setting, and the right part is the video frame outputs of the AVSS setting. The other two methods only produce segmentation maps that are not that precise, whereas our AVSAC can not only significantly evade some false alarms but also more accurately delineate the shapes of sounding objects.}\label{vis}
\end{figure*}

\section{Experiments}
\label{others}

\subsection{Datasets}

\textbf{AVSBench-Object} \cite{zhou2022audio} is an audio-visual segmentation dataset with pixel-level annotations. Each video is 5 seconds long and is uniformly segmented into five clips. Annotations for the sounding objects are provided by the binary mask of the final frame of each video clip. The dataset includes two subsets according to the number of audio sources in a video: a semi-supervised single sound source subset (S4), and a fully supervised multi-source subset (MS3). The S4 subset consists of 4932 videos, while 
the MS3 subset contains 424 videos. We follow the dataset split of \cite{zhou2022audio} for training, validation, and testing.

\textbf{AVSBench-Semantic} \cite{zhou2023audio} is an extension of the AVSBench-Object dataset with 12,356 videos across 70 categories. It is designed for audio-visual semantic segmentation (AVSS). Different from AVSBench-Object which only has binary mask annotations, AVSBench-Semantic offers semantic-level annotations for sounding objects. The videos in AVSBench-Semantic are longer, each lasting 10 seconds, and 10 frames are extracted from each video for prediction. In terms of volume, the AVSBench-Semantic dataset has grown to nearly triple the size of the original AVSBench-Object dataset, featuring 8,498 videos for training, 1,304 for validation, and 1,554 for testing.

%\textbf{AVSS} \citep{zhou2023audio} is an extension of the AVSBench-Object dataset with 12,356 videos across 70 categories. It is designed for audio-visual semantic segmentation (AVSS). Different from the AVSBench-Object dataset which only has binary mask annotations, AVSS offers semantic-level annotations for sounding objects. The videos in AVSS are longer, each lasting 10 seconds, and 10 frames are extracted from each video for prediction. 

\subsection{Evaluation Metrics}
We follow \cite{zhou2022audio} and employ Jaccard index (the mean Intersection-over-Union between the predicted masks and the ground truth) and F-score as evaluation metrics.

\subsection{Implementation  Details}
We train our AVSAC model for the three sub-tasks using 2 NVIDIA A6000 GPUs. We freeze the parameters of visual and audio backbones. We choose the Deformable Transformer \cite{zhu2020deformable} block as the structure of the self-attention layer in the vision-guided decoder. Consistent with previous works, we choose AdamW \cite{loshchilov2017decoupled} as our optimizer, with the batch size set to 4 and the initial learning rate set to $2\times 10^{-5}$. All video frames are resized to $224 \times 224$ resolution. Since the MS3 subset is small in scale, we train it for 60 epochs, while for the large-scale S4 and AVSS subsets, we train them for 30 epochs. The encoder and decoder in our AVSAC both are comprised of 6 layers with an embedding dimension of 256. We provide our detailed experimental settings used for each sub-task in the appendix for reference.%We set $\lambda_{f}$, $\lambda_{d}$ and $\lambda_{avfs}$ to 1, 1 and 1 for the best performance. 

\subsection{Qualitative  Comparison}
We compare the qualitative results between AVSBench \cite{zhou2022audio}, AVSegFormer \cite{gao2023avsegformer}, and our AVSAC in Figure~\ref{vis}. PVT-v2 is adopted as the visual encoder backbone of all three methods, among which our AVSAC enjoys the finest shape-aware segmentation effect and significantly relieves some false alarms. The left side of Figure~\ref{vis} visualizes the fineness of our method in delineating the shape of the sounding object. The middle part of Figure~\ref{vis} visualizes that our AVSAC can curtail some false alarms. Specifically, when the dog appears in the video frames but does not make any sound, AVSBench cannot accurately find the correct-sounding objects but chooses to include all of them. AVSegFormer performs better but still mistakes some objects in the background as the sounding source. In contrast, our AVSAC can accurately segment the sounding baby without generating false alarm predictions. On the right side, AVSegFormer even completely misses the sounding man in two frames, while AVSBench confounds the pigeon with the sounding object and makes errors. The reason for the flaws of the other two methods (also other AVS methods) is that they do not realize their modality fusion mode actually deteriorates the audio-visual imbalance degree, since they overwhelm the weak audio cues with the dominant visual features so that not enough audio guidance can be given to distinguish the correct-sounding object. Our method relieves modality imbalance and balances visual representations with enough audio cues by building bidirectional bridges between our two-tower decoders.

\begin{figure}
    \centering   \includegraphics[width=0.98\textwidth,height=0.32\textwidth]{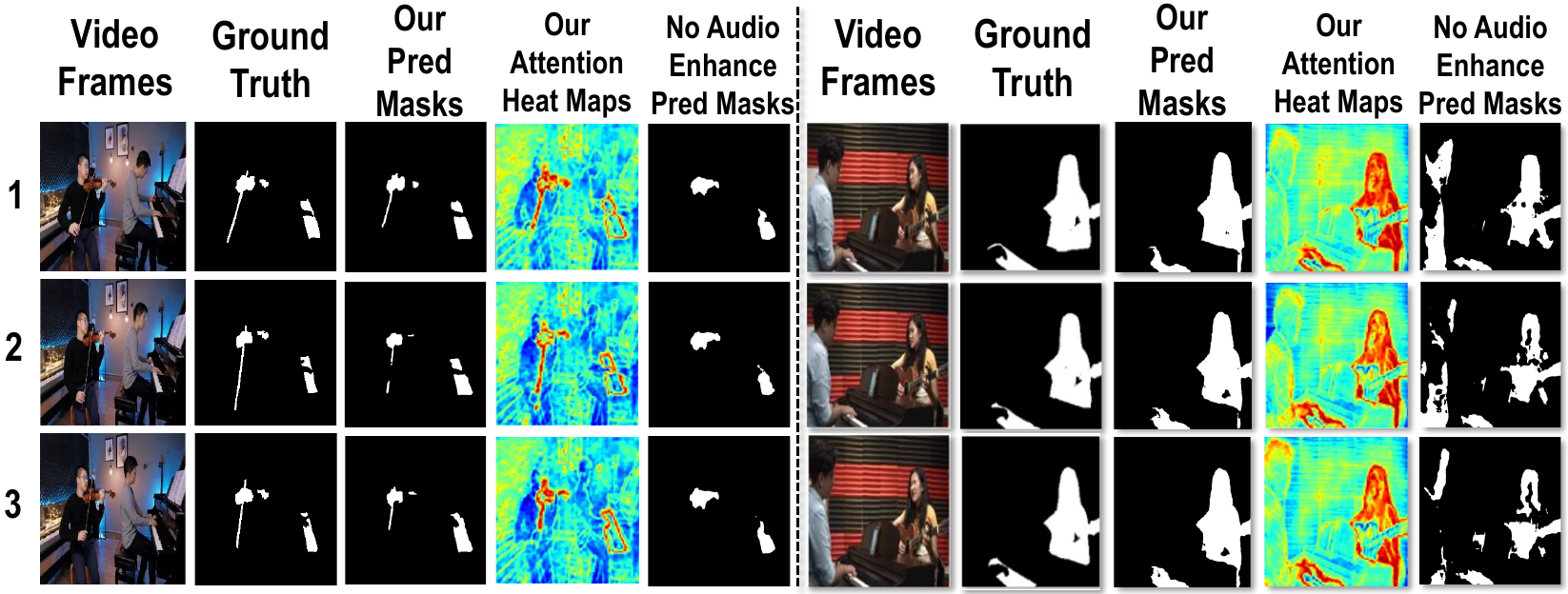}
    \caption{Visual comparison of AVS results over consecutive frames with and without multiple bidirectional bridges. The red areas in attention heat maps highlight the sounding objects.}\label{att_vis}
\end{figure}

We also visualize the attention maps of consecutive frames with and without audio enhancement in Figure~\ref{att_vis} to further prove the effectiveness of our method. Obviously, no-audio-enhancement results are coarser than audio-enhanced ones from the left part of Figure~\ref{att_vis}. The right part of Figure~\ref{att_vis} again visualizes the false alarms caused by modality imbalance--the left man is mistaken as a sounding object even though he does not make any sound. Equipped with our bidirectional bridges between two decoders for continuous audio enhancement can significantly mitigate this issue. %For more qualitative results, please refer to our appendix.

\subsection{Quantitative Comparison}
The quantitative results on the test set of S4, MS3, and AVSS are presented in Table~\ref{tab_instance}, where our method has achieved SOTA performances on all three AVS subsets in terms of F-score and mIoU. There is a consistent performance gain of our AVSAC compared to other methods, regardless of whether ResNet-50 \cite{he2016deep}, Swin-Transformer \cite{liu2021swin} or PVT2 \cite{wang2022pvt} is used as the backbone. This suggests that the modality imbalance problem indeed exists and limits the performance of present AVS methods. By building multiple bidirectional bridges between two paralleled decoders, we balance audio and visual modality information flows during the whole training process, therefore solving the modality imbalance issue and significantly boosting the segmentation performances. Note that the performance improvement benefits from our two-tower decoder structure with bidirectional bridges itself under our AVFS guidance, instead of coming from more trainable parameters. Actually, the parameter number of our baseline AVSegFormer \cite{gao2023avsegformer} is 186M, while our model is slightly more lightweight with 181M parameters, but our model far outperforms present AVS methods.

\begin{table*}
\caption{Comparison with other SOTA methods on the AVS benchmark. The methods are evaluated on three AVS sub-tasks, including single sound source segmentation (S4), multiple sound source segmentation (MS3), and audio-visual semantic segmentation (AVSS). The evaluation metrics are F-score and mIoU. The higher the better. The figures in bold mark the highest ones in each column.}
\vskip 0.15in
\begin{center}
\begin{small}
\begin{sc}
\centering
\begin{center}
\setlength{\tabcolsep}{0.8mm}
\begin{tabular}{ l | c | c | c | c | c | c | c}
		\toprule[1pt]
		\multirow{2}{*}{Methods} & \multirow{2}{*}{Backbone} & \multicolumn{2}{c|}{S4} & \multicolumn{2}{c|}{MS3} & \multicolumn{2}{c}{AVSS}\\
		\cline{3-8}        
		       & & F-score & mIoU & F-score & mIoU & F-score & mIoU\\
		\midrule
  LVS \cite{chen2021localizing}& Res-50 \cite{he2016deep}& 51.0 & 37.94 & 33.0 & 29.45 &--&--\\
  MSSL \cite{qian2020multiple}& Res-18 \cite{he2016deep}& 66.3 & 44.89 & 36.3 & 26.13 &--&--\\
  3DC \cite{mahadevan2020making}& Res-34 \cite{he2016deep}& 75.9 & 57.10 & 50.3 & 36.92 &21.6& 17.27\\
  SST \cite{duke2021sstvos}& Res-101 \cite{he2016deep}& 80.1 & 66.29 & 57.2 & 42.57 &--&--\\
  AOT \cite{yang2021associating}& Swin-B \cite{liu2021swin}& 82.8 & 74.2 & 0.050 & 0.018 &31.0& 25.40\\
  iGAN \cite{mao2021transformer}& Swin-T \cite{liu2021swin}& 77.8 & 61.59 & 54.4 & 42.89 &--&--\\
  LGVT \cite{zhang2021learning}& Swin-T \cite{liu2021swin}& 87.3 & 74.94 & 59.3 & 40.71 &--&--\\
  \midrule
  AVSBench-R50 \cite{zhou2022audio}& Res-50 \cite{he2016deep}& 84.8 & 72.79 & 57.8 & 47.88 &25.2 & 20.18\\
  AVSegFormer-R50 \cite{gao2023avsegformer}& Res-50 \cite{he2016deep}& 85.9 & 76.45 & 62.8 & 49.53 &29.3& 24.93\\
  AVSBG-R50 \cite{hao2023improving} & Res-50 \cite{he2016deep}& 85.4 & 74.13 & 56.8 & 44.95 &--&--\\
  AuTR-R50 \cite{liu2023audio} & Res-50 \cite{he2016deep}& 85.2 & 75.0 & 61.2 & 49.4 &--&--\\
  DiffAVS-R50 \cite{mao2023contrastive} & Res-50 \cite{he2016deep}& 86.9 & 75.80 & 62.1 & 49.77 &--&--\\
 %AQFormer-R50 \cite{huang2023discovering} & Res-50 \cite{he2016deep}& 86.4 & \textbf{77.0} & \textbf{66.9} & \textbf{55.7} & &\\
 CATR-R50 \cite{li2023catr} & Res-50 \cite{he2016deep}& \textbf{87.1} &74.9 &65.6 &53.1 & --&--\\
 AVSAC-R50 (Ours) & Res-50 \cite{he2016deep}& 86.95 & \textbf{76.90} & \textbf{65.81} & \textbf{53.95} &\textbf{29.71}& \textbf{25.43}\\%151.46M

  \midrule
  AVSBench-PVT \cite{zhou2022audio}& PVT2 \cite{wang2022pvt}& 87.9 & 78.74 & 64.5 & 54.00 &35.2& 29.77\\
  AVSegFormer-PVT \cite{gao2023avsegformer}& PVT2 \cite{wang2022pvt}& 89.9 & 82.06 & 69.3 & 58.36 &42.0& 36.66\\
  AVSBG-PVT \cite{hao2023improving} & PVT2 \cite{wang2022pvt}& 90.4 & 81.71 & 66.8 & 55.10 &--&--\\
  AuTR-PVT \cite{liu2023audio} & PVT2 \cite{wang2022pvt}& 89.1 & 80.4 & 67.2 & 56.2 &--&--\\
  DiffAVS-PVT \cite{mao2023contrastive} & PVT2 \cite{wang2022pvt}& 90.2 & 81.38 & 70.9 & 58.18 &--&--\\
  %AQFormer-PVT \cite{huang2023discovering} & PVT2 \cite{wang2022pvt}& 89.4 & 81.6 & 72.1 & 61.1 & &\\
  CATR-PVT \cite{li2023catr}& PVT2 \cite{wang2022pvt}& 91.3 &84.4 & 76.5 &62.7 &38.5 &32.8\\
  AVSAC-PVT (Ours) & PVT2 \cite{wang2022pvt}& \textbf{91.56} & \textbf{84.51} & \textbf{76.60} & \textbf{64.15} &\textbf{42.39}& \textbf{36.98}\\
  \bottomrule[1pt]
	\end{tabular}
\label{tab_instance}
	\end{center}
\end{sc}
\end{small}
\end{center}
\vskip -0.1in
\end{table*}

\section{Ablation Studies}
In this section, we conduct ablation experiments to verify the effectiveness of each key design in our proposed AVSAC. Specifically, we adopt PVTv2 \citep{wang2022pvt} as the backbone and conduct extensive experiments on the S4 and MS3 sub-tasks.

\textbf{Ablation of each key module.} In Table~\ref{ab1}, we ablate the influence of each of the key components within our framework. We adopt AVSegFormer \cite{gao2023avsegformer} as the baseline. We can discern from Table~\ref{ab1} that compared with our AVFS, our BAVD yields a more pronounced performance gain, especially in the MS3 subset, which poses greater challenges. The improvement can be attributed to BAVD's bidirectional bridge mechanism, which consistently strengthens the audio cues throughout the decoding process to guarantee that the audio cues will not fade away. In this way, the previously dominant visual features can be balanced by strengthened audio signals, which helps the network to more accurately delineate sounding object shapes according to the strengthened audio guidance. Our AVFS is more like an auxiliary module to enable the visual feature to learn audio cues in a frame-wise manner. This approach furnishes our BAVD with fine-grained guidance, furthering our efforts to mitigate the imbalance between modalities.

\begin{table}
\caption{Ablation study of each key design. Results show that BAVD and AVFS complements each other and combining both performs better.}
\vskip 0.15in
\begin{center}
\begin{small}
\begin{sc}
		\centering
		\label{tab:ablation}
		\setlength{\tabcolsep}{0.5mm}
		%\vspace{-5mm}
		
		\begin{tabular}{l| c | c | c | c | r }
    \toprule
        \multirow{2}{*}{No.} & \multirow{2}{*}{Settings} & \multicolumn{2}{c|}{S4} & \multicolumn{2}{c}{MS3}\\
        \cline{3-6}
        & &F-score & mIoU & F-score & mIoU\\
        \midrule
    $\#$1& Baseline  &  89.9 & 82.06 & 69.3 & 58.36\\
    %$\#$2& Audio $\rightarrow$ Vis  & 90.64  & 83.28 & 72.86&  61.98\\
    %$\#$3& Vis $\rightarrow$ Audio  &  90.12  & 82.46 & 70.03 & 58.89 \\   
    $\#$2& BAVD  &  90.93  & 83.65 & 75.12&  63.50\\
    $\#$3& Ours & \textbf{91.56} & \textbf{84.51} & \textbf{76.60} & \textbf{64.15}\\
        \bottomrule
    \end{tabular}\label{ab1}
	
\end{sc}
\end{small}
\end{center}
\vskip -0.1in
\end{table}

%To ablate the effectiveness of our BAVD, we adopt our  (denoted by IAVI* in Model $\#$2) by replacing the cross attention output in Figure~\ref{module} (a) with the audio feature $F_{audio}$. 

\textbf{Ablation of BAVD.} The ablation studies of the internal structure of BAVD are presented in Table~\ref{ab2}. In this work, we propose BAVD, which is a two-tower decoder structure with bidirectional bridges to enhance audio cues to reach a continuous audio-visual modality balance and interaction. To demonstrate the necessity of bidirectional bridges, we compare this design with two kinds of unidirectional bridges--the audio-to-vision bridge (Model $\#1$) and the vision-to-audio bridge (Model $\#2$). The audio-to-vision bridge brings higher performance gain than the vision-to-audio bridge, which proves that strengthening audio cues as visual guidance is very effective in relieving modality imbalance. Remarkably, the implementation of bidirectional bridges achieves the highest performance, corroborating the efficacy of this approach in reaching a harmonious audio-visual balance. We also ablate the necessity of the AGCA blocks in the AGV decoder branch (Model $\#3$) and observe a performance decline when the AGCA in the AGV decoder is deleted, so we keep it. The reason for the drop is that not enough audio guidance gets injected into the visual decoder branch.

\begin{table}
\caption{Ablation study of BAVD. It shows that building bidirectional bridges between two-tower decoders improves more.}
\vskip 0.15in
\begin{center}
\begin{small}
\begin{sc}
		\centering
		\label{tab:ablation2}
		\setlength{\tabcolsep}{0.5 mm}
		
		\begin{tabular}{c| c | c | c | c | c }
    \toprule
        \multirow{2}{*}{No.} & \multirow{2}{*}{Settings} & \multicolumn{2}{c|}{S4} & \multicolumn{2}{c}{MS3}\\
        \cline{3-6}
       & &F-score & mIoU & F-score & mIoU\\
        \midrule
    $\#$1& only Audio $\rightarrow$ Vis & 90.70  & 83.32 & 72.91&  62.05\\
    $\#$2& only Vis $\rightarrow$ Audio & 90.16  & 82.49 & 71.02 & 59.35 \\
    $\#$3& w/o AGCA &  90.86  & 83.51 & 73.33 & 62.67\\
    %$\#$4& +SA in VGAD &  91.01  & 83.86 &73.66 & 62.99\\
    $\#$4& w/ BAVD & \textbf{91.56} & \textbf{84.51} & \textbf{76.60} & \textbf{64.15}\\
        \bottomrule
    \end{tabular}\label{ab2}
\end{sc}
\end{small}
\end{center}
\vskip -0.1in
\end{table}

%We also ablate the necessity of attention blocks in both decoder branches by deleting each AGCA in AGV decoder branch (Model $\#3$) and adding a self-attention between each VGA decoder branch layers (Model $\#4$). %Results show that adding self-attention between each VGA layer slightly harms AVS fineness because the audio feature itself is not as information-rich as visual features, so processing it with excessive parameters may cause ambiguity, so we do not add a self-attention block to the audio decoder. 

\textbf{Ablation of the loss function.} We ablate the impact of the loss function in Table~\ref{ab3}. The results of the first two rows indicate that combining Focal loss with Dice loss as the segmentation loss contributes better to segmentation fineness, because the Dice loss solves the foreground-background imbalance problem but ignores a further imbalance between easy and difficult samples, whereas Focal loss can focus on these difficult misclassified samples. The results of the last three rows demonstrate that equipping the segmentation loss with our proposed AVFS loss brings the best results. This is because our AVFS loss facilitates a frame-wise alignment of audio cues with visual features, empowering the visual features to draw from audio cues in a fine-grained manner, thereby more increasing the proportion of audio components in the visual features. Also, our experimental results suggest that KL-divergence is more suitable than L2-Norm in quantifying the per-frame divergence between audio and visual modalities. The efficacy of KL-divergence stems from its non-symmetric characteristic, allowing it to discern whether a distribution functions as the reference model or the approximation. This is particularly advantageous when we need to approximate an audio distribution with a visual one. %prevent the network from learning some harmful data bias and also help reconstruct some lost audio cues to guarantee modality balance.

%This is because our AVFS loss facilitates a frame-wise alignment of audio cues with visual features, empowering the audio cues to draw from visual signals for further enhancement. Moreover, our experimental results suggest that KL-divergence is more adept than L2-Norm at quantifying the divergence between audio and visual modalities. The asymmetric nature of KL-divergence, which uniquely identifies the distribution serving as the reference model from the one being approximated, is especially advantageous when the goal is to model audio distributions using visual counterparts.

\begin{table}
\caption{Ablation study of loss function. Combining the segmentation loss (Focal+Dice) with our KL-divergence-based AVFS loss works the best.}
\centering
\label{tab:ablation4}
\vskip 0.15in
\begin{center}
\begin{small}
\begin{sc}
\setlength{\tabcolsep}{0.5mm}
		%\vspace{-5mm}
		\begin{tabular}{c | c | c | c | c }
    \toprule
        \multirow {2}{*}{Settings} & \multicolumn{2}{c|}{S4} & \multicolumn{2}{c}{MS3}\\
        \cline{2-5}
        & F-score & mIoU & F-score & mIoU\\
        \midrule
    Dice Loss & 90.56  & 83.29 & 73.66& 62.99\\
   $\mathcal{L}_{Seg}$ &  90.93  & 83.65 & 75.12&  63.50\\
    %w/o $\mathcal{L}_{S3}$ &  90.93  & 83.65 & 74.12&  63.46\\
    %$\mathcal{L}_{Seg}$ + $\mathcal{L}_{AVTS}$ $\rightarrow$ cos &    &  & &  \\
    $\mathcal{L}_{Seg}$ + L2 & 91.02   & 83.80 & 75.56 & 63.57\\
    $\mathcal{L}_{Seg}$ + Frame-wise L2 &  91.15  & 83.94 & 75.78 & 63.66\\
    Ours & \textbf{91.56} & \textbf{84.51} & \textbf{76.60} & \textbf{64.15}\\
        \bottomrule
    \end{tabular}\label{ab3}
	
\end{sc}
\end{small}
\end{center}
\vskip -0.1in
\end{table}

\textbf{Ablation of the number of decoder layers.} We further explore the impact of varying the number of BAVD layers, denoted by $N$, and present our findings in Table~\ref{ab4}. As we increment the number of BAVD layers from 2 to 6, a consistent gain in AVS performance is observed. Optimal results on both subsets are achieved when $N$ is set to 6, suggesting that more BAVD layers contribute to an enhanced performance. This can be attributed to the fact that with an insufficient number of decoder layers, audio cues cannot receive adequate enhancement, thus failing to effectively mitigate the audio-visual modality imbalance issue and resulting in comparatively modest performance improvements.

\begin{table}
\caption{Ablation study of decoder layer number. We find that 6 decoder layers work better than other configurations.}
\centering
\vskip 0.15in
\begin{center}
\begin{small}
\begin{sc}
\setlength{\tabcolsep}{0.5mm}
		%\vspace{-5mm}
		\begin{tabular}{c | c | c | c | c }
    \toprule
        \multirow{2}{*}{Layers} & \multicolumn{2}{c|}{S4} & \multicolumn{2}{c}{MS3}\\
        \cline{2-5}
        &F-score & mIoU & F-score & mIoU\\
        \midrule
    2  &  88.76  & 79.83 & 71.58& 59.83\\
    4 &  90.50  & 82.84 & 73.56& 62.88 \\
    6 & \textbf{91.56} & \textbf{84.51} & \textbf{76.60} & \textbf{64.15}\\
        \bottomrule
    \end{tabular}\label{ab4}
	
\end{sc}
\end{small}
\end{center}
\vskip -0.1in
\end{table}

\section{Conclusion}
In this paper, we notice that present AVS methods are plagued by a modality imbalance issue: visual features take the dominance and overshadow the audio cues, shrinking the effect of audio guidance in the AVS task. This invariably restricts the overall performance of AVS. To solve this, we try to bootstrap the AVS task by strengthening the audio cues, thereby achieving a more equitable interplay between the audio and visual modalities. As a result, we propose the AVSAC network. The superior performances on all three AVS datasets demonstrate the effectiveness of our method in solving modality imbalance in AVS. Our work not only mitigates the modality imbalance but also paves the way for future innovations in creating more harmonious and effective audio-visual alignments and interactions.

\textbf{Limitations.} Similar to other methods that adopt Transformers, our proposed method does not exhibit significant advantages in model size and inference efficiency compared to recent AVS methods.

%\subsubsection*{Acknowledgments}
%We would like to acknowledge SentEval \cite{conneau2018senteval} authors for making the code open-source and freely available. We are thankful to Roberto Silveira for the GPU time donation to execute all the experiments. 

%\bibliography{references.bib}
\bibliography{arXiv/example_paper.bib}

\appendix
\section{Appendix}
\subsection{Supplemental Results}
We list the detailed settings of our model in Table~\ref{setting}.
\begin{table}[H]
\caption{Detailed  settings.  This  table   provides   a   detailed 
overview  of  the specific  settings used  for  each  sub-task.} \label{table_complex}
\vskip 0.15in
\begin{center}
\begin{small}
\begin{sc}
\centering
    \begin{tabular}{cccc}
    \toprule
        Settings & S4  & MS3 & AVSS\\
        %\cline{2-9}
        \hline
        %\hline
        input resolution  & 224 $\times$ 224& 224 $\times$ 224 & 224 $\times$ 224\\
        frames T  & 5& 5& 10\\
        embedding dimension D & 256 & 256 & 256\\
        AGV decoder feature size  & 5376 $\times$ 256 & 5376 $\times$ 256 & 1029 $\times$ 256\\
        VGA decoder feature size  & 300 $\times$ 256 & 300 $\times$ 256 & 300 $\times$ 256\\
        decoder layers N  & 6 & 6 & 6\\
        batch size  & 4 & 4 & 4\\
        optimizer & AdamW & AdamW & AdamW\\
        learning rate & $2\times 10^{-5}$ & $2\times 10^{-5}$ & $2\times 10^{-5}$\\
        drop path rate & 0.1 & 0.1 & 0.1\\
        epoch & 30 & 60 & 30\\
        \bottomrule
    \end{tabular}\label{setting}
\end{sc}
\end{small}
\end{center}
\vskip -0.1in
\end{table}

\end{document}